\documentclass[runningheads]{llncs}
\usepackage[T1]{fontenc}
\usepackage{ulem}
\usepackage{multirow}
\usepackage{array}
\usepackage{makecell}

\usepackage{array}
\usepackage{fontawesome}
\usepackage[dvipsnames]{xcolor}

\newcolumntype{C}[1]{>{\centering\arraybackslash}p{#1}}

\newcommand{\yequals}{\textcolor{Goldenrod}{\faArrowsH}}
\newcommand{\greenup}{\textcolor{ForestGreen}{\faAngleUp}}
\newcommand{\reddown}{\textcolor{red}{\faAngleDown}}

\usepackage[english]{babel}

\usepackage{amsmath}
\usepackage{graphicx}
\usepackage[colorlinks=true, allcolors=blue]{hyperref}

\begin{document}

\title{ICDAR 2023 Competition on Robust Layout Segmentation in Corporate Documents}
\author{Christoph Auer, Ahmed Nassar, Maksym Lysak, Michele Dolfi, Nikolaos Livathinos, Peter Staar}

\author{Christoph Auer\orcidID{0000-0001-5761-0422} \and
Ahmed Nassar\orcidID{0000-0002-9468-0822} \and
Maksym Lysak\orcidID{0000-0002-3723-6960}  \and
Michele Dolfi\orcidID{0000-0001-7216-8505}  \and
Nikolaos Livathinos\orcidID{0000-0001-8513-3491} \and
Peter Staar\orcidID{0000-0002-8088-0823}}

\titlerunning{ICDAR 2023 Comp. on Robust Layout Segmentation in Corporate Docs}
\authorrunning{C. Auer, et al.}
%
\institute{IBM Research\\
\email{\{cau,ahn,mly,dol,nli,taa\}@zurich.ibm.com}}
\maketitle              
\begin{abstract}

Transforming documents into machine-processable representations is a challenging task due to their complex structures and variability in formats. Recovering the layout structure and content from PDF files or scanned material has remained a key problem for decades. ICDAR has a long tradition in hosting competitions to benchmark the state-of-the-art and encourage the development of novel solutions to document layout understanding. In this report, we present the results of our \textit{ICDAR 2023 Competition on Robust Layout Segmentation in Corporate Documents}, which posed the challenge to accurately segment the page layout in a broad range of document styles and domains, including corporate reports, technical literature and patents. To raise the bar over previous competitions, we engineered a hard competition dataset and proposed the recent DocLayNet dataset for training. We recorded 45 team registrations and received official submissions from 21 teams. In the presented solutions, we recognize interesting combinations of recent computer vision models, data augmentation strategies and ensemble methods to achieve remarkable accuracy in the task we posed. A clear trend towards adoption of vision-transformer based methods is evident. The results demonstrate substantial progress towards achieving robust and highly generalizing methods for document layout understanding.

\keywords{Document Layout Analysis \and Machine Learning  \and Computer Vision \and Object Detection \and ICDAR Competition}
\end{abstract}

\section{Introduction}
Document understanding is a key business process in the data-driven economy since documents are central to knowledge discovery and business insights. Converting documents into a machine-processable format is a particular challenge due to their huge variability in formats and complex structure. Recovering the layout structure and content from either PDF files or scanned material has remained a key problem since decades, and is as relevant-as-ever today. One can find vast amounts of approaches and solutions to this task \cite{huang2022layoutlmv3,li2022dit,CCS,newccs:pagemodelrnn,TableFormer,DocLayNet}, all of which are constrained to different degrees in the domains and document styles that they can perform well on. A highly generalising model for structure and layout understanding has yet to be achieved. 

ICDAR has organized various competitions in the past to benchmark the state-of-the-art and encourage the development of novel approaches and solutions to layout segmentation problems in documents~\cite{ICDAR2013,ICDAR2017,ICDAR2019,Jimeno-etal-2021-PubLayNetCompetition}. In this report, we present the results of our \textit{ICDAR 2023 Competition on Robust Layout Segmentation in Corporate Documents}, which posed the challenge to accurately segment the layout of a broad range of document styles and domains, including corporate reports, technical literature and patents. Participants were challenged to develop a method that could identify layout components in document pages as bounding boxes. These components include paragraphs, (sub)titles, tables, figures, lists, mathematical formulas, and several more. The performance of submissions was evaluated using the commonplace mean average precision metric (mAP) used in the COCO object detection competition~\cite{lin2015microsoft}. To raise the bar over previous competitions, we proposed to use our recently published DocLayNet dataset~\cite{DocLayNet} for model training, and engineered a challenging, multi-modal competition dataset with a unique distribution of new page samples.

Below, we present a detailed overview of this competition, including its datasets, evaluation metrics, participation, and results.

\section{Datasets}

\begin{figure}
\centering
\includegraphics[width=1\textwidth]{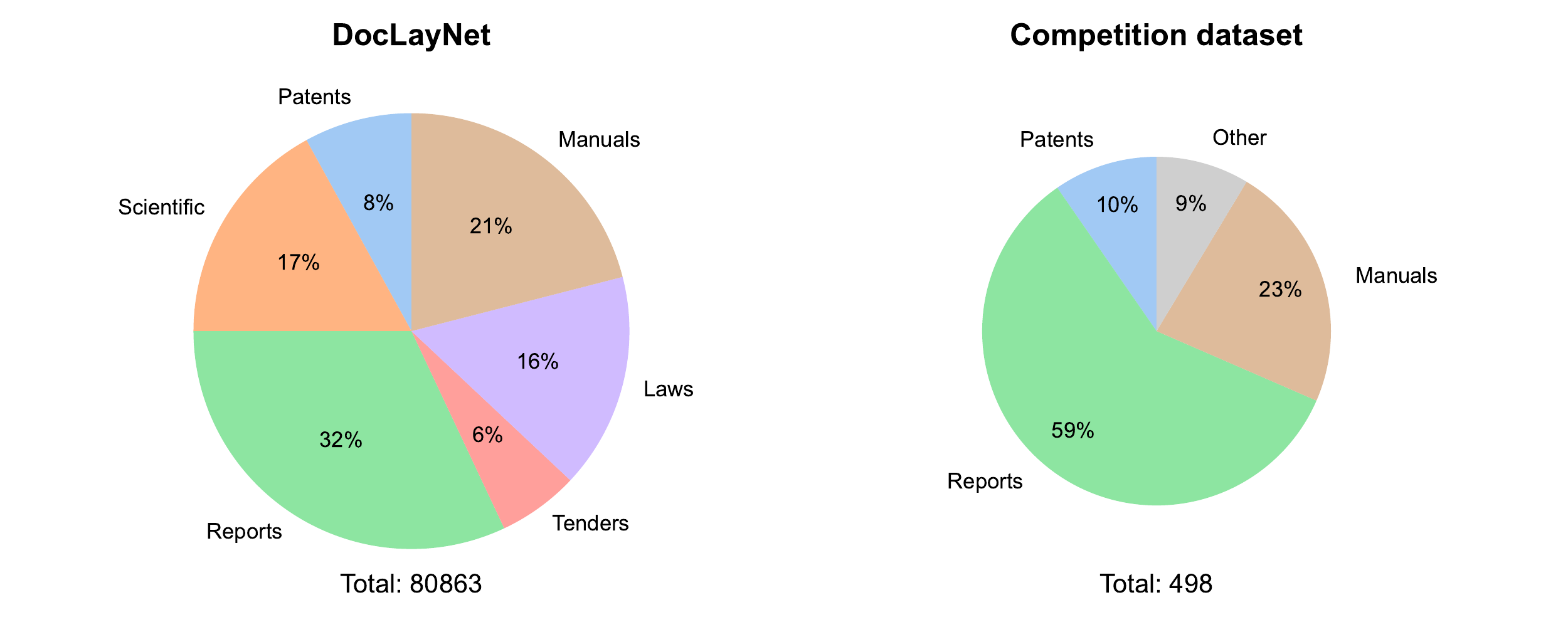}
\caption{\label{fig:dataset_stats}Dataset statistics of DocLayNet and the competition dataset.}
\end{figure}

\subsection{Related Work}

Layout segmentation datasets published in the recent past, such as PubLayNet~\cite{zhong2019publaynet} or DocBank~\cite{Li-etal-2020-docbank}, have enabled a big leap forward for ML-driven document understanding approaches due to their huge ground-truth size compared to earlier work. However, these datasets still remain limited to a narrow domain of predominantly scientific documents, which is owed to their automatic ground-truth generation approach from mostly uniform XML or \LaTeX{} sources. Despite exposing many different publisher layouts, all documents strongly share common traits and general structure. This has led to a saturation of ML model accuracy baselines at a very high level, with little room for improvement~\cite{Jimeno-etal-2021-PubLayNetCompetition}. Yet, all publicly proposed ML models trained on these datasets generalize rather poorly to out-of-domain document samples, such as those found in the corporate world. For example, tables in invoices or manuals are difficult to detect correctly with models trained on scientific literature or books.

\begin{figure}[b!]
\centering
\includegraphics[width=1\textwidth]{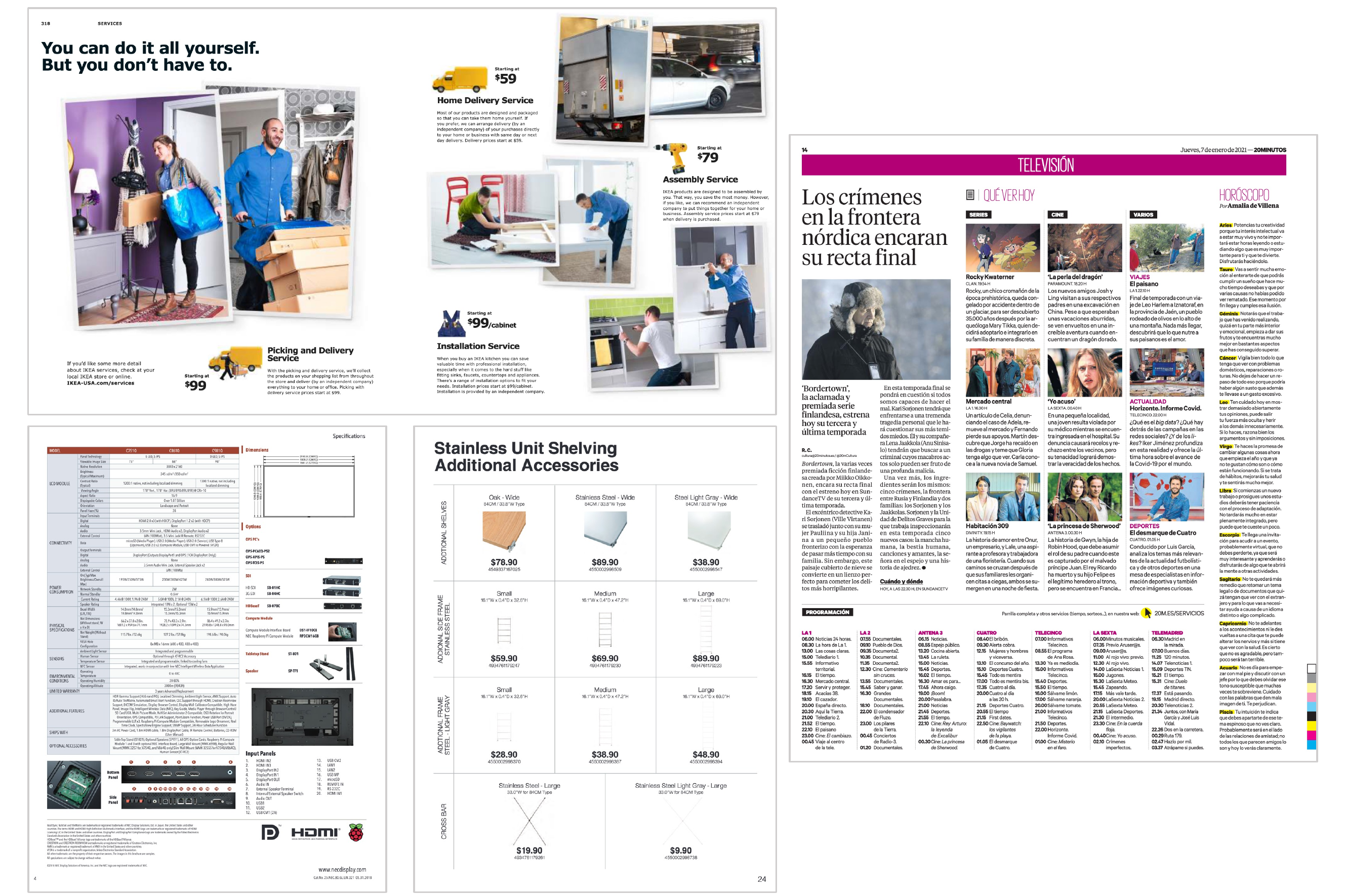}
\caption{\label{fig:examples}Select samples in the competition dataset (\textit{Other} category) which fall outside of the layout distribution in DocLayNet.}
\end{figure}

\subsection{DocLayNet dataset}

The DocLayNet dataset~\cite{DocLayNet} addresses these known limitations by providing 80,863 page samples from a broad range of document styles and domains, which are fully layout annotated by human experts to a high-quality standard. DocLayNet is the first large-scale dataset covering a wide range of layout styles and domains, which includes Financial reports, Patents, Manuals, Laws, Tenders, and Technical Papers. It defines 11 class labels for rectangular bounding-box annotations, namely \textit{Caption}, \textit{Footnote}, \textit{Formula}, \textit{List-item}, \textit{Page-footer}, \textit{Page-header}, \textit{Picture}, \textit{Section-header}, \textit{Table}, \textit{Text} and \textit{Title}. Detailed instructions and guidance on how to consistently annotate the layout of DocLayNet pages were published in the accompanying layout annotation guideline.

Additionally, DocLayNet provides a JSON representation of each page with the original text tokens and coordinates from the programmatic PDF code. This opens the opportunity for new multi-modal ML approaches to the layout segmentation problem.

\subsection{Competition dataset}
To assess the layout segmentation performance of each team's submissions, we engineered a competition dataset of 498 new pages in the same representation as the original DocLayNet dataset, which was provided to the participants without any annotation ground-truth. This competition dataset includes a mix of corporate document samples as shown in Fig.~\ref{fig:examples}. Samples in the new \textit{Other} category expose layouts which fall outside of the DocLayNet layout space.

\section{Task}

\begin{figure}[ht!]
\centering
\includegraphics[width=0.75\textwidth]{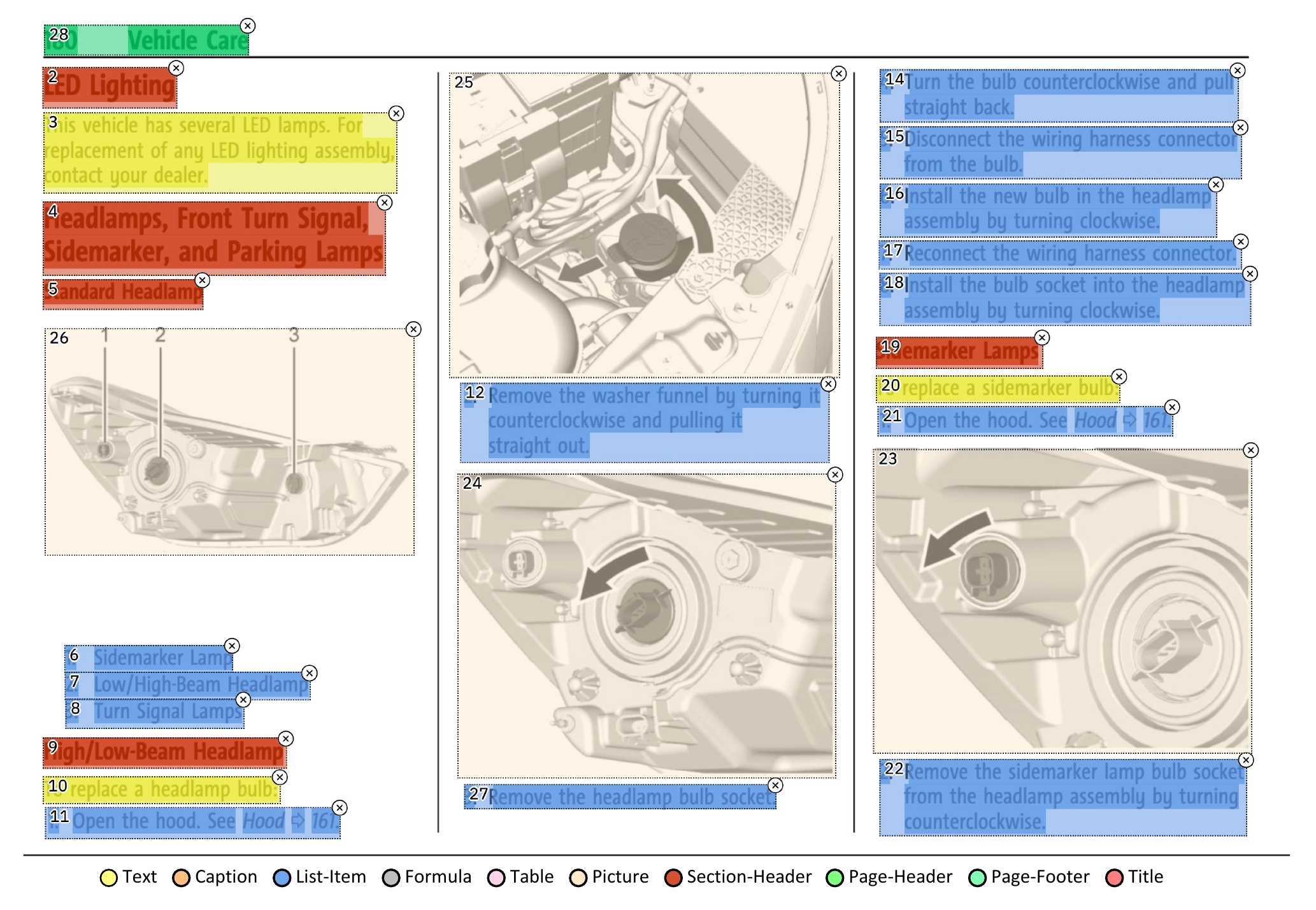}
\caption{\label{fig:annotation-example}Example page with bounding-box annotations. }
\end{figure}

We designed the competition objective as a straightforward object detection task, since this is well-understood in the computer-vision community and fits the representation format of our DocLayNet dataset. Participants of our competition were challenged to develop methods that can identify layout components in document pages as rectangular bounding boxes, labelled with one of the 11 classes defined in the DocLayNet dataset (see Fig.~\ref{fig:annotation-example}). The performance of each team's approach was evaluated on our competition dataset using the well established COCO mAP metric.

\subsubsection*{Submission format:}
Since the COCO dataset format~\cite{lin2015microsoft} and tooling is well established in the object detection community, we provided a standard COCO dataset file as part of our competition dataset, which includes the definition of class labels and image identifiers, but no ground-truth annotation data. Submissions were expected in the format of a JSON file complying with the commonly used \textit{COCO results} schema, including complete bounding-box predictions for each page sample, matching to the identifiers defined in our provided dataset file.

\subsubsection*{Evaluation metric:}

All submissions were evaluated using the Mean Average Precision (mAP) @ Intersection-over-Union (IoU) [0.50:0.95] metric, as used in the COCO object detection competition. In detail, we calculate the average precision for a sequence of IoU thresholds ranging from 0.50 to 0.95 with a step size of 0.05, using the standard \textit{pycocotools} library\footnote{\href{https://pypi.org/project/pycocotools/}{pypi.org/project/pycocotools}}. This metric was computed for every document category in the competition dataset separately. Then, the mean of the mAPs across all categories was computed with equal weights per category. The final ranking of every team's submissions was based on the overall mAP.

\section{Competition}

\subsubsection*{Schedule:}

Our competition was officially announced on December 19th, 2022 and ended on April 3rd, 2023. The regular competition phase ended on March 26th, 2023 and the final week was run as a dedicated extension phase. Results of both phases are reflected in section \ref{sec:results}.

\subsubsection*{Setup:}

We launched a competition website\footnote{\href{https://ds4sd.github.io/icdar23-doclaynet}{ds4sd.github.io/icdar23-doclaynet}} to provide task descriptions, instructions, resources and news updates for the competition. For submission management, automatic online evaluation and tracking team submissions on a leader board, we relied on the free-to-use \textit{EvalAI} platform~\cite{yadav2019evalai}. To ensure fair conditions and prevent reverse engineering of our ground-truth, each team was originally granted 10 submission attempts on the evaluation platform. We increased this limit by 5 attempts for the extension phase. The feature in EvalAI to declare submissions private or public allowed teams to create multiple private submissions and check how they perform in evaluation before deciding to re-submit one of them as an official entry. The test-score for each submission was provided directly after submission. The latter has advantages and disadvantages. On the one hand, teams have a direct feedback on the quality of their results and can explore different strategies, which is one of the main motivations of this competition. On the other hand, it can also be used to overfit the model. For this explicit reason, we limited the number of submissions of each team to 10 (with extension 15). To set a baseline for the leader board, the competition organizers created an initial submission entry, which was visible to all teams.

\section{\label{sec:results}Results}

\subsection{Overview}

After the competition ended, we counted 45 team registrations, which altogether created 374 private or public submissions. Out of these, 21 team decided to make at least one public submission which counts towards the final ranking.
Table \ref{tab:leaderboard} shows the results achieved by the participating teams for the regular submission phase and the extension phase of the competition. More detailed analysis and descriptions of selected methods from the participants are presented below.

\begin{table}[!ht]
    \centering
    \caption{\label{tab:leaderboard}Leaderboard of our competition with all teams ranking above our baseline (rank 19). Ranks are shown separately for the regular phase (\textit{reg}) and the extension phase (\textit{ext}}
    \small
    \begin{tabular}{p{4.2cm}|lllll|p{1cm}p{1cm}l}
    \textbf{}                                                     & \multicolumn{5}{c|}{\textbf{mAPs after extension}}                     & \multicolumn{3}{c}{\textbf{Ranking}}                     \\
    \textbf{Team}                                                 & \textbf{Overall} & \textbf{Rep} & \textbf{Man} & \textbf{Pat} & \textbf{Other} & \textbf{reg.} & \textbf{ext.} & \textbf{Diff}  \\ 
    \hline
    \textbf{docdog}          & \textbf{0.70}  & \textbf{0.66}    & \textbf{0.69}    & 0.84             & \textbf{0.62}  & 1                & 1                  & \yequals \\
    \textbf{BOE\_AIoT\_CTO}                      & 0.64           & 0.54             & 0.67             & 0.84             & 0.52           & 2                & 2                  & \yequals \\
    \textbf{INNERCONV}                           & 0.63           & 0.57             & 0.63             & \textbf{0.85}    & 0.48           & 3                & 3                  & \yequals \\
    \textbf{LC-OCR}                              & 0.63           & 0.61             & 0.65             & 0.77             & 0.48           & 5                & 4                  & \greenup 1 \\
    \textbf{DXM-DI-AI-CV-TEAM}                   & 0.63           & 0.54             & 0.63             & 0.82             & 0.51           & 4                & 5                  & \reddown 1 \\
    \textbf{alexsue}                             & 0.61           & 0.53             & 0.63             & 0.81             & 0.49           & 6                & 6                  & \yequals \\
    \textbf{PIX}                                 & 0.61           & 0.52             & 0.63             & 0.82             & 0.46           & -                & 7                  & * \\
    \textbf{Acodis}                              & 0.60           & 0.53             & 0.62             & 0.80             & 0.46           & 15               & 8                  & \greenup 7 \\
    \textbf{Linkus}                              & 0.59           & 0.49             & 0.64             & 0.77             & 0.48           & 7                & 9                  & \reddown 2 \\
    \textbf{TTW}                                 & 0.58           & 0.49             & 0.61             & 0.80             & 0.42           & 8                & 10                 & \reddown 2 \\
    \textbf{amdoc}                               & 0.58           & 0.47             & 0.62             & 0.80             & 0.42           & 12               & 11                 & \greenup 1 \\
    \textbf{CVC-DAG}                             & 0.58           & 0.49             & 0.61             & 0.77             & 0.44           & 9                & 12                 & \reddown 3 \\
    \textbf{SPDB LAB}                            & 0.57           & 0.48             & 0.57             & 0.80             & 0.44           & 10               & 13                 & \reddown 3 \\
    \textbf{Alphastream.ai}                      & 0.57           & 0.47             & 0.57             & 0.79             & 0.45           & 11               & 14                 & \reddown 3 \\
    \textbf{Hisign}                              & 0.57           & 0.48             & 0.62             & 0.79             & 0.39           & 13               & 15                 & \reddown 2 \\
    \textbf{DLVC}                                & 0.55           & 0.53             & 0.57             & 0.74             & 0.38           & 14               & 16                 & \reddown 2 \\
    \textbf{Vamshikancharla}                     & 0.49           & 0.36             & 0.48             & 0.76             & 0.37           & 16               & 17                 & \reddown 1 \\
    \textbf{Azure}                               & 0.49           & 0.44             & 0.55             & 0.59             & 0.38           & 17               & 18                 & \reddown 1 \\
    \textbf{ICDAR23 DocLayNet\newline organizers (Baseline)} & 0.49           & 0.38             & 0.52             & 0.70             & 0.35           & 18               & 19                 & \reddown 1               
    \end{tabular}

\end{table}

\subsection{General analysis}

\subsubsection*{Layout segmentation performance:}

\begin{figure}[htb!]
\centering
\includegraphics[width=1\textwidth]{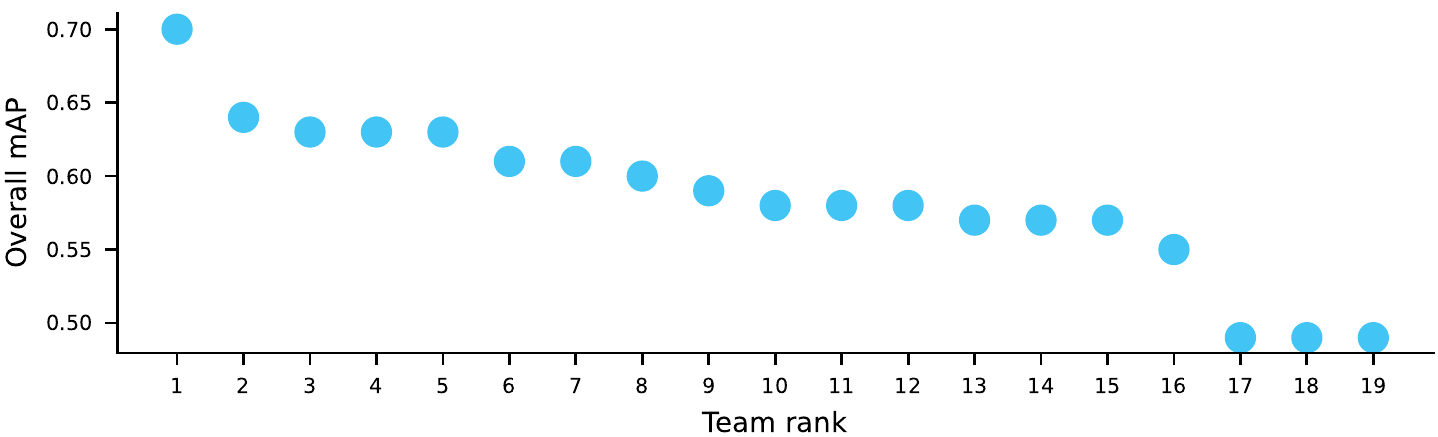}
\caption{\label{fig:distmap}Distribution of overall mAP achieved by teams. Numbers and ranks refer to extension phase. }
\end{figure}

It is apparent that the top-ranking team (docdog) has presented a solution that is performing notably superior compared to the remainder of the field, as evidenced by their 6\% lead in total mAP score over the second best submission. This result is achieved through outperforming every other team in the \textit{Reports} category (5\% lead) and the particularly difficult \textit{Others} category (10\% lead). From the second rank down, we observe a very competitive field with many teams achieving similar levels of mAP performance, ranging from 0.64 (rank 2) to 0.55 (rank 16). Two more teams ranked just slightly above our baseline mAP of 0.49 (see Fig.~\ref{fig:distmap}).

Throughout the extension phase, we observed mostly small improvements of overall mAP within a 1-2\% range, with few exceptions such as team \textit{docdog} and team \textit{Acodis}, which managed to improve by 4\% and 6\% over their result from the regular competition phase, respectively. Team \textit{PIX} joined as a new entrant in the extension phase only.

The highest, and also most consistent performance across submissions, is observed in the \textit{Patent} category, with 12 teams achieving an mAP of 0.79 or better. This is consistent with our expectations, since Patent document layouts are the most uniform and structured. The diverse, free-style layouts in the \textit{Reports} and \textit{Others} categories posed a considerably bigger challenge, with mAPs generally ranging in the low 60\%s and 50\%s respectively.

Two interesting observations can be made. On one hand, we find significantly lower mAPs in the submissions across our competition set categories than those which were achieved for example on PubMed Central papers in the ICDAR 2021 Competition on Scientific Literature Parsing\cite{Jimeno-etal-2021-PubLayNetCompetition}. This can be attributed both to the more challenging layouts and higher class count of the DocLayNet dataset, as well as to the distribution bias and hard samples we engineered the competition dataset to expose.
On the other hand, we see a significant spread of mAPs across the final submissions, with almost all teams exceeding the baseline by a significant margin. This delivers evidence that the participating teams have created solutions that differentiate themselves significantly from previous off-the-shelf object detection methods (see baseline). It also shows that the investment to develop sophisticated methods is beneficial to obtain superior performance on this dataset.

\subsubsection*{Models and Strategies:}

In the solutions presented by the top five teams, we were pleased to see novel and interesting combinations of recent computer vision models, data augmentation strategies and ensemble methods applied to solve the layout segmentation task with high accuracy. All top-ranking solutions adopt, to different degrees, the recently emerging deep-learning models based on vision transformer methods and self-supervised pre-training, such as the generic DINO\cite{zhang2022dino} and MaskDINO\cite{li2022mask} models, or the document-understanding focused DiT\cite{li2022dit} and LayoutLMv3\cite{huang2022layoutlmv3} models. The DocLayNet dataset was used for fine-tuning in this context. Several solutions  combine these new-generation vision models with more traditional, CNN-based object detectors such as YOLO~\cite{Jocher_YOLO_by_Ultralytics_2023} through model ensemble, for example through \textit{Weighted Boxes Fusion}~\cite{solovyev2021weighted}. 
Data augmentation strategies used by the teams include multi-scale and mosaic methods, as well as deriving synthetic datasets from DocLayNet. Two of the top five teams reported that they include the additional text cell layer provided by DocLayNet and the competition dataset in their approach. No teams stated to create private ground-truth data that was not derived from DocLayNet.

\subsection{Method descriptions}

Below we summarize the methods reported by the top five teams for comparison reasons to our best understanding. We would like to extend our thanks to all competition teams who took the time to provide us with a comprehensive description of their methods.

\subsection*{Team docdog (Tencent WeChat AI)}

The team created a synthetic image dataset of 300,000 samples based on the training dataset. For the task of layout prediction, the team used two models, YOLOv8~\cite{Jocher_YOLO_by_Ultralytics_2023} and DINO~\cite{zhang2022dino}. An extra classification model was trained to categorize the samples of the competition dataset into the document categories. 
YOLOv8 models with different network sizes (medium, large, x-large) were trained, each with different input resolutions, for ensemble and optimization of the detection performance.
For the DINO model, the team applied a carefully designed augmentation strategy and integrated focal modulation networks~\cite{NEURIPS2022_focal_modulation_networks} in the backbone for improved performance. Separate models were trained per category, both with and without synthetic data.
Model hyper-parameters were optimized using a Tree-Structured Parzen Estimator (TPE)~\cite{NIPS2011_86e8f7ab} to find the best weights.
Prediction results from the individual models were combined using Weighted Boxes Fusion (WBF)~\cite{solovyev2021weighted} and fine-tuned using text cell coordinates from the JSON representation of the samples in the competition dataset.
Further detail on the approach is provided in the team's \textit{WeLayout} paper~\cite{zhang2023welayout}.

\subsection*{Team BOE\_AIoT\_CTO}

The team relied exclusively on the DocLayNet dataset for training, and applied scale and mosaic methods for image augmentation. For the task of layout prediction, the team trained two object detection models, YOLOv5~\cite{glenn_jocher_2022_7347926} and YOLOv8~\cite{Jocher_YOLO_by_Ultralytics_2023}.
Training was conducted over 150 epochs using BCELoss with FocalLoss, and mosaic augmentation was cancelled for the final 20 epochs. Additionally, a DiT model~\cite{li2022dit} (dit-large) was fine-tuned using the DocLayNet dataset. To improve vertical text detection, the team added multi-scale image training.
Predictions for the final submission were ensembled from three detectors to achieve superior performance.

\subsection*{Team INNERCONV}

For the task of layout prediction, the team uses the MaskDINO model~\cite{li2022mask}. MaskDINO is derivative of DINO~\cite{zhang2022dino} which introduces a mask prediction branch in parallel to the box prediction branch of DINO. It achieves better alignment of features between detection and segmentation. In training, only the image representation of the DocLayNet dataset is used. In inference, the team applied the Weighted Boxes Fusion (WBF) technique~\cite{solovyev2021weighted} to ensemble the predictions on multiple scales of the same input image.

\subsection*{Team LC-OCR (CVTE)}

For the task of layout prediction, the team applied two models, VSR~\cite{zhang2021vsr} and LayoutLMv3~\cite{huang2022layoutlmv3}, which use pre-trained weights. Prediction results from both models are merged in inference. Detections for the classes \textit{Footnote}, \textit{Picture}, \textit{Table} and \textit{Title} were taken from LayoutLMv3, the remainder of classes from VSR. In VSR, the team included the text cell information provided in the JSON representation of DocLayNet.

\subsection*{Team DXM-DI-AI-CV-TEAM (Du Xiaoman Financial)}

For the task of layout prediction, the team trained different versions of Cascade Mask R-CNN~\cite{he2018mask} models, based on a DiT\cite{li2022dit} backbone (DiT-large), and fuse prediction results using different models.

\subsection*{Baseline of ICDAR 2023 DocLayNet organizers}

To set a comparison baseline for the competition, the organizers used a YOLOv5 model (medium size), and trained it solely on the DocLayNet training dataset, with images re-scaled to square 1024 by 1024 pixels. The model was trained from scratch with default settings for 80 epochs. We applied standard augmentation techniques such as mosaic, scale, flipping, rotation, mix-up and image levels. 

\section{Conclusions}

We believe that this ICDAR competition served its purpose well to benchmark the state-of-the-art solutions to the layout segmentation task in documents, and again encouraged the development of unique new approaches.
Our new competition dataset was designed to raise the bar over previous competitions by providing diverse, challenging page layouts, paired with multi-modal representation. This enabled participants to test the generalization power of the latest computer-vision methods, especially with recently emerging models based on self-supervised pre-training and visual transformers.

We were pleasantly surprised by the high level of engagement in this competition, with 45 teams registering, out of which 21 teams created an official final submission. The budget of 15 total submissions was fully used by the majority of the contestants. Overall, the level of sophistication demonstrated in the approaches went well beyond our anticipation. One core take-away is the importance of data augmentation and ensemble techniques to improve the layout prediction performance beyond the level of what any single end-to-end model currently delivers. It was also interesting to observe how the various techniques applied by the different teams in many cases yielded similar results in overall accuracy.
The remarkable progress demonstrated by the top-performing teams in this competition will be valuable for future research on highly capable document understanding models.

We are also glad to see this competition spark wider interest in the community, as it prompted some members to build and share fully runnable example codes and publish blog articles on training and inference with DocLayNet and pre-trained models\cite{guillou_2023}. To support these community efforts, we made DocLayNet available on the HuggingFace datasets hub\footnote{https://huggingface.co/datasets/ds4sd/DocLayNet}. As such, we believe that this ICDAR competition has also helped to establish the DocLayNet dataset as a well known asset for document understanding research and applications.  

\subsection*{Acknowledgements}
We would like to thank all participants for their remarkable efforts and contributions to this competition, and the Competitions Chairs for providing the opportunity to host this competition in ICDAR 2023.

\bibliographystyle{splncs04}
\bibliography{sample}

\end{document}